\pdfoutput=1

\documentclass[11pt]{article}

\usepackage[final]{acl}

\usepackage{times}
\usepackage{latexsym}

\usepackage[T1]{fontenc}

\usepackage[utf8]{inputenc}

\usepackage{microtype}

\usepackage{algorithm}
\usepackage{algorithmic}
\usepackage{amsfonts}
\usepackage{amsthm} 
\usepackage{amsmath}
\usepackage{array}
\usepackage{booktabs}
\usepackage{comment}
\usepackage{colortbl}
\usepackage{hyperref}
\usepackage{graphicx}
\usepackage{multirow} 
\usepackage{subfigure}
\usepackage{subcaption}
\usepackage{url}
\usepackage{wrapfig}
\usepackage{xcolor}

\title{LVPruning: An Effective yet Simple Language-Guided Vision Token Pruning Approach for Multi-modal Large Language Models}

\author{
 \textbf{Yizheng Sun\textsuperscript{1}},
 \textbf{Yanze Xin\textsuperscript{2}},
 \textbf{Hao Li\textsuperscript{1}},
 \\
 \textbf{Jingyuan Sun\textsuperscript{1,*}},
 \textbf{Chenghua Lin\textsuperscript{1}},
 \textbf{Riza Batista-Navarro\textsuperscript{1}},
\\
\\
 \textsuperscript{1}University of Manchester,
 \textsuperscript{2}Imperial College London,
\\
 \small{
   \textbf{*Correspondence:} \href{jingyuan.sun@manchester.ac.uk}{jingyuan.sun@manchester.ac.uk}
 }
}

\begin{document}

\maketitle


\begin{abstract}
Multi-modal Large Language Models (MLLMs) have achieved remarkable success by integrating visual and textual modalities. However, they incur significant computational overhead due to the large number of vision tokens processed, limiting their practicality in resource-constrained environments. We introduce Language-Guided Vision Token Pruning (LVPruning) for MLLMs, an effective yet simple method that significantly reduces the computational burden while preserving model performance. LVPruning employs cross-attention modules to compute the importance of vision tokens based on their interaction with language tokens, determining which to prune. Importantly, LVPruning can be integrated without modifying the original MLLM parameters, which makes LVPruning simple to apply or remove. Our experiments show that LVPruning can effectively reduce up to 90\% of vision tokens by the middle layer of LLaVA-1.5, resulting in a 62.1\% decrease in inference Tera Floating-Point Operations Per Second (TFLOPs), with an average performance loss of just 0.45\% across nine multi-modal benchmarks.

\end{abstract}







\section{Introduction}
Multi-modal Large Language Models (MLLMs) have achieved impressive results by combining visual and textual information to perform complex tasks that require understanding both modalities \citep{DBLP:conf/nips/Dai0LTZW0FH23, DBLP:conf/icml/0008LSH23,DBLP:journals/corr/abs-2310-03744, DBLP:conf/haii/SunLLB24}. However, these models can be highly computationally intensive, limiting their practicality in resource-constrained environments \citep{DBLP:journals/corr/abs-2310-03744, DBLP:conf/nips/LiuLWL23a}. One important fact that leads to such substantial computational overhead is that these models often process a large number of vision tokens representing input image patches, but not all visual information is equally important for understanding. The human brain, for instance, can focus on salient features while ignoring irrelevant details, allowing for highly efficient visual perception \citep{doi:10.1080/02724988843000104}. Inspired by this, there is a growing need to develop MLLMs that can prioritize crucial vision tokens, reducing computational costs without largely sacrificing performance.

\begin{figure}[t]
\centering
        \includegraphics[scale=0.36]{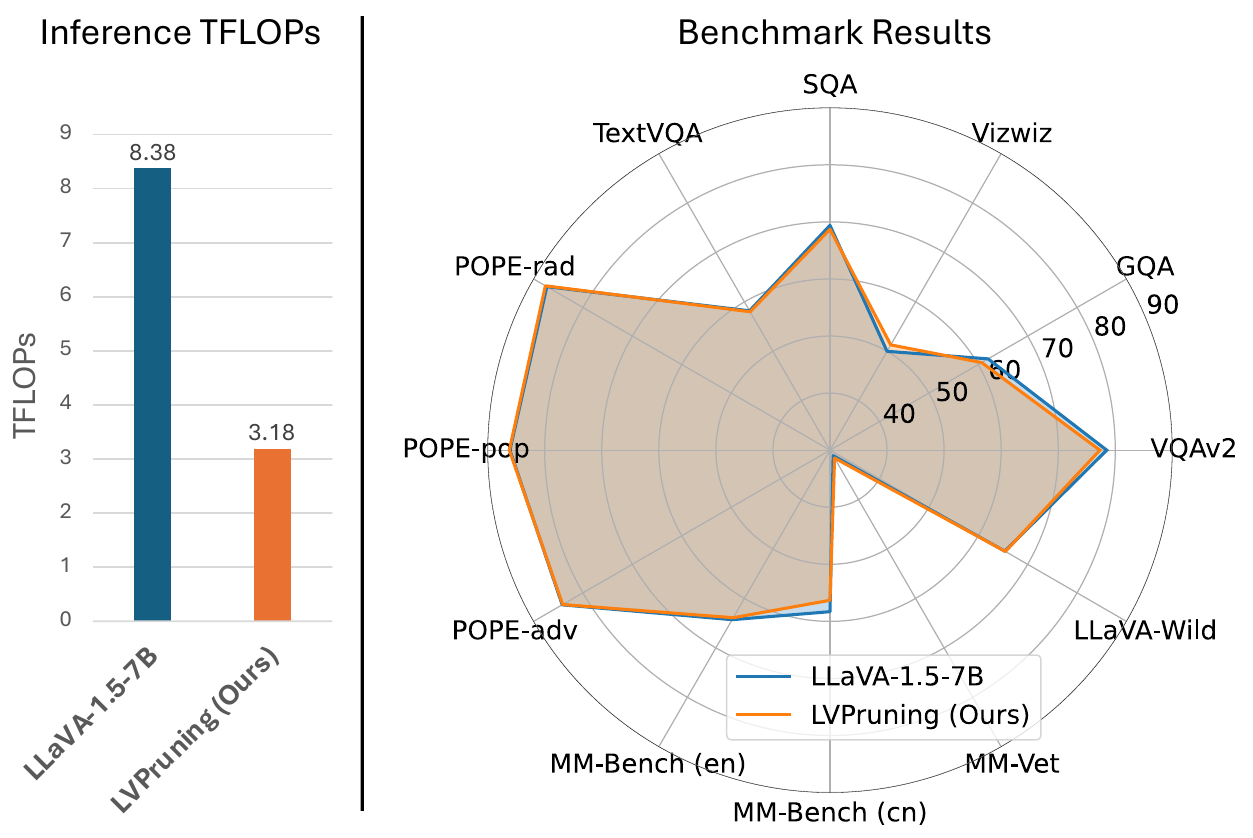}
        \caption{LVPruning can reduce 62.1\% of inference TFLOPs for LLaVA-1.5-7B with marginal performance loss across nine multi-modal benchmarks. *All TFLOPs reported in this paper are computed using a dummy input consisting of 1 image and 30 text tokens.} \label{fig:intro_rada}
\end{figure}

Previous approaches to enhance the computational efficiency of MLLMs have explored various strategies. Models utilizing Q-former as the vision encoder condense visual information into a smaller set of tokens. Though effectively reducing computational load, such condensation potentially leads to a loss of essential visual information, compromising performance \citep{DBLP:conf/icml/0008LSH23, DBLP:conf/nips/Dai0LTZW0FH23, DBLP:conf/iclr/Zhu0SLE24}. On the other hand, models such as LLaVA pass all vision tokens through a simple Multi-Layer Perceptron (MLP) connector to the language model, achieving high performance but at the cost of increased computational demands \citep{DBLP:journals/corr/abs-2310-03744, DBLP:conf/nips/LiuLWL23a}. Additionally, token compression techniques \citep{DBLP:conf/nips/RaoZLLZH21,DBLP:conf/iclr/BolyaFDZFH23,DBLP:conf/iccv/ChenSXLZCJQL23} that detect important vision tokens solely based on visual features have shown promise in single-modal tasks but can not make full of the interaction between visual and linguistic information in MLLMs. These highlight a trade-off between computational efficiency and model performance, indicating a need for solutions that can balance both aspects effectively and efficiently.


To address these challenges, we propose Language-Guided Vision Token Pruning (LVPruning), a simple yet effective method that dynamically reduces the number of vision tokens in MLLMs based on their relevance to the language context. We introduce lightweight cross-attention decision modules where vision tokens attend to language tokens to compute importance scores. This relevance scoring allows the model to decide whether to keep or prune each vision token, effectively filtering out less informative visual data. By integrating these decision modules into various layers of the MLLM, LVPruning enables progressive token pruning as the model processes deeper layers. During training, we freeze all original model parameters and only train the inserted decision modules, ensuring that the base model remains unchanged and the pruning mechanism can be easily applied or removed.


Our contributions are threefold. First, as shown in Figure \ref{fig:intro_rada}, we demonstrate that LVPruning can significantly reduce computational costs—up to a 62.1\% decrease in inference TFLOPs—by pruning as much as 90\% of vision tokens without substantially affecting model performance. Second, we introduce a novel, language-guided token pruning mechanism that is both effective and easy to integrate into existing MLLMs, requiring minimal changes to the original architecture. Third, our method allows for adjustable token pruning ratios during inference without retraining, offering flexibility in balancing efficiency and performance according to specific needs. Through extensive experiments on various multi-modal benchmarks, we show that LVPruning provides a practical solution to enhance the efficiency of MLLMs while maintaining their ability to understand and generate accurate multi-modal content.

\section{Related Work}
\textbf{Multi-modal Large Language Models}: Recent advancements in MLLMs have significantly enhanced the integration of visual and textual modalities. BLIP-2 \citep{DBLP:conf/icml/0008LSH23} introduced a two-stage learning framework that connects pre-trained vision models with language models using a Q-former as a vision encoder, effectively generating a condensed set of vision tokens for efficient processing. Building on it, InstructBLIP \citep{DBLP:conf/nips/Dai0LTZW0FH23} and MiniGPT-4 \citep{DBLP:conf/iclr/Zhu0SLE24} incorporated instruction tuning to improve the model's ability to follow complex prompts and perform diverse tasks. Alternatively, models such as LLaVA-1.5 \citep{DBLP:journals/corr/abs-2310-03744} directly input all vision tokens from pre-trained vision encoders into the language model. While this approach achieves higher performance owing to richer visual information, it results in substantial computational overhead. These models exemplify the trade-off between computational efficiency and performance, underscoring the need for approaches that can balance both aspects without compromising accuracy.

\textbf{Efficient Transformers}: Many techniques have been proposed to improve computation efficiency for Transformer models, such as knowledge distillation \citep{DBLP:journals/corr/HintonVD15}, token merging/pruning \citep{DBLP:conf/iclr/BolyaFDZFH23, DBLP:conf/nips/RaoZLLZH21, DBLP:conf/iccv/ChenSXLZCJQL23}, and quantization \citep{DBLP:journals/corr/GongLYB14, DBLP:conf/cvpr/WangLLLH19}. For NLP tasks, methods like DistillBERT \citep{DBLP:journals/corr/abs-1910-01108}, MiniLM \citep{DBLP:conf/nips/WangW0B0020} use knowledge distillation to create smaller models for more efficient inference. For computer vision tasks, \citet{DBLP:journals/corr/abs-2202-07800, DBLP:conf/iclr/BolyaFDZFH23} and \citet{DBLP:conf/iccv/ChenSXLZCJQL23} focus on pruning or merging tokens based on their importance in image classification. These approaches reduce the number of vision tokens by identifying less informative patches or merging similar tokens during inference. The closest work to ours is DynamicVit \citep{DBLP:conf/nips/RaoZLLZH21}, who use MLP layers to predict token pruning decisions. They hierarchically insert multiple pruning layers into Vision-Transformer-based models for the image classification task. However, these methods are specifically designed for single-modal targets and do not address challenges in multi-modal settings. Our research distinguishes itself by focusing on token pruning for MLLMs in the context of image comprehension tasks.

\begin{figure}[t]
\centering
    \includegraphics[scale=0.85]{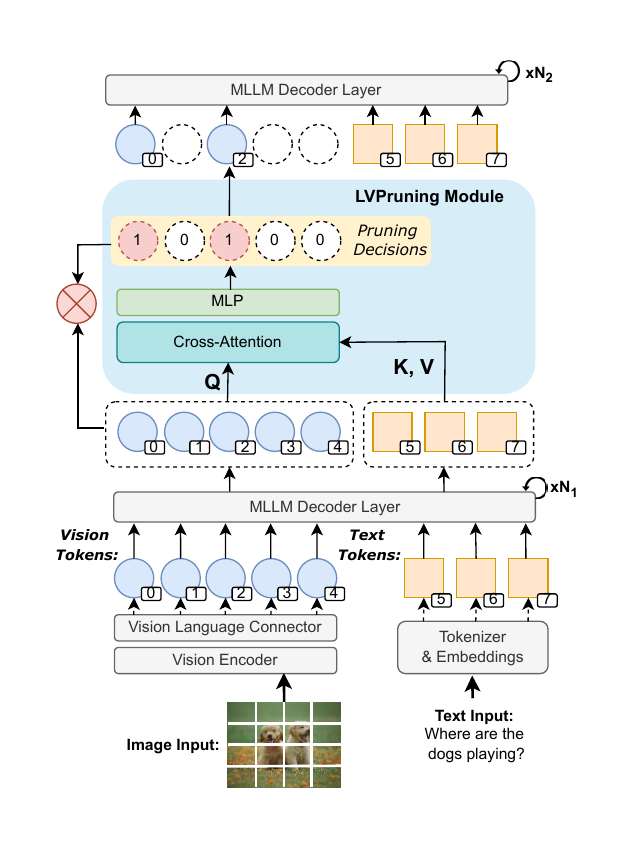}
    \caption{Overall Framework Architecture. LVPruning modules are incorporated into specific layers of an MLLM, where vision tokens serve as queries and language tokens act as keys and values. A pruning decision is predicted for each vision token. The operation denoted by $\otimes$ applies these decisions—serving as attention masking during training and token removal via indexing during inference.}
    \label{fig:framework_arch}
\end{figure}

\section{Methodology}

In this section, we present the LVPruning framework in detail. As shown in Figure \ref{fig:framework_arch}, LVPruning is designed for MLLMs that pass all vision tokens through an MLP connector into the language model. The architecture consists of a Transformer-based pre-trained CLIP vision encoder, an MLP vision-language connector, and an LLM backbone. First, an image input is divided into patches and processed by the CLIP model \citep{DBLP:conf/icml/RadfordKHRGASAM21}, such that each image patch becomes a representative vision token. Next, with the vision language connector projecting vision tokens into the dimension of the LLM's text space, the concatenated vision and text tokens are fed into the LLM for causal text generation. In specific layers of the LLM, cross-attention decision modules dynamically select the most salient token (the one with the highest attention score) to guide inference, removing redundant tokens. During training, we apply attention masks \citep{DBLP:conf/nips/RaoZLLZH21} to mask out pruned vision tokens. Importantly, instead of updating positional embeddings after token pruning, we retain the original positional embeddings for the remaining vision tokens, as used in standard LLMs \citep{DBLP:journals/corr/abs-2307-09288}.

\subsection{Cross-Attention Decision Module}

We now describe the detailed architecture of the cross-attention decision module designed for token pruning. A decision module, which is responsible for selecting and discarding vision tokens, comprises cross-attention layers and an MLP layer. Multiple instances of these modules are inserted into different layers of the LLM backbone for progressive pruning. Let $\mathbf{H} \in \mathbb{R}^{N \times d}$ represent the output from an LLM hidden layer, where $\boldsymbol{N}$ is the sequence length and $\boldsymbol{d}$ is the dimension of the hidden representations. $\mathbf{H}$ contains the subset of vision tokens and text tokens. We define the set of the vision tokens indices as
$\mathbf{I_V}=\{n_{v_i} \mid  v_i \in \mathbb{N}, 0 \leq v_i < N\}$,
and the set of text tokens indices as
$\mathbf{I_T}=\{n_{t_i} \mid  t_i \in \mathbb{N}, (0 < t_i \leq N) \wedge (t_i \notin I_V)\}$. We use the vision tokens as the query tokens
\begin{equation} \label{equ1}
    \mathbf{Q} = W_qH_{I_V} \in \mathbb{R}^{\mid I_V \mid \times d},
\end{equation}
and text tokens as the Key and Value tokens
\begin{equation} \label{equ2}
    \mathbf{K,V} = (W_{k}/W_{v})H_{I_T} \in \mathbb{R}^{\mid I_T \mid \times d},
\end{equation}
where $W_q, W_k, W_v$ are linear projection layers. We then compute the attention matrix and feed the output to an FFN, as described by \citet{DBLP:conf/nips/VaswaniSPUJGKP17}.
\begin{equation} \label{equ3}
    \mathbf{O} = \text{Softmax}\left(\frac{QK^T}{\sqrt{d}}\right)V + Q.
\end{equation}
Inspired by \citet{DBLP:conf/nips/RaoZLLZH21}, we feed the output from the FFN to a linear layer $W_{O}$ to predict the scores of keeping and removing a vision token:
\begin{equation} \label{equ4}
    \boldsymbol{\gamma} = W_{O} O \in \mathbb{R}^{\mid I_V \mid \times 2},
\end{equation}
where $\boldsymbol{\gamma_{i, 0}}$ represents the score for keeping the vision token $\mathbf{H_{{I_{V,i}}}}$ and $\boldsymbol{\gamma_{i, 1}}$ represents the score for removing the vision token $\mathbf{H_{{I_{V,i}}}}$. The decisions of vision token pruning are then generated based on $\boldsymbol{\gamma}$. The mechanism for generating and applying decisions differs between training and inference. Multiple decision modules are inserted into different layers of the LLM, such that the vision tokens are pruned progressively throughout the LLM.

\subsection{End-to-End Training}
To ensure that the process of generating and applying token pruning decisions based on $\boldsymbol{\gamma}$ is differentiable, where $\boldsymbol{\gamma}$ denoted as the output from each cross-attention decision module, we draw inspiration from \citet{DBLP:conf/nips/RaoZLLZH21}. Specifically, we apply the Gumbel-Softmax distribution to $\boldsymbol{\gamma}$, redistributing it into one-hot vectors $\boldsymbol{D^{GS}}$. The first dimension of each vector is then used as the decision $\boldsymbol{D}$, determining whether to retain a given token:
\begin{equation} \label{equ6}
    \boldsymbol{D^{GS}}=\text{Gumbel-Softmax}(\gamma) \in \{0,1\} ^ {\mid I_V \mid \times 2},
\end{equation}
\begin{equation} \label{equ7}
    \boldsymbol{D}=D^{GS}_{:,0} \in \{0,1\} ^ {\mid I_V \mid \times 1},
\end{equation}
where $\boldsymbol{D_i = 1}$ means we keep the vision token $\mathbf{H_{{I_{V,i}}}}$ and vice versa. The number of kept tokens given by the decision is not fixed during training, and directly removing unwanted vision tokens will impede batch processing. To solve this concern, we make attention masks $\boldsymbol{M}$ based on $\boldsymbol{D}$ for both vision and language tokens. Specifically, 
\begin{multline}\label{equ8}
    \boldsymbol{M_{i,j}} = 
    \begin{cases} 
    1 & \text{if } i = j \text{ or } j \in I_T\\
    D_j & \text{if } i \neq j  \text{ and } j \in I_V
    \end{cases}, \\ \quad 1 \leq i, j \leq {N}.
\end{multline}
Therefore, $\boldsymbol{M}$ is constructed such that all language tokens are assigned a mask value of 1, while the vision tokens are masked according to the token pruning decisions $\boldsymbol{D}$. Additionally, all diagonal elements of $\boldsymbol{M}$ are set to 1 to improve numerical stability. 

However, $\boldsymbol{M}$ disregards the original causal and padding attention masks. To address this, we first apply the original attention mask $\boldsymbol{\bar{M}}$ to the raw attention scores to obtain causal attention matrix $\boldsymbol{\bar{A}}$. Then, we apply $\boldsymbol{M}$ with the $Softmax$ operation to $\boldsymbol{\bar{A}}$ to get the final attention matrix $\boldsymbol{\hat{A}}$. Specifically, we define $\boldsymbol{M_l}$ as the attention mask generated from the decision module at layer $\boldsymbol{l}$. The attention scores at layer $\boldsymbol{l+x}$ is calculated by
\begin{equation}\label{equ12}
    \boldsymbol{\hat{A}_{l+x}} = \hat{Softmax}(\bar{A}_{l+x}, M_l),
\end{equation}
\begin{multline}\label{equ13}
    \hat{Softmax}(A, M) = \frac{\exp(A_{i,j}) M_{i,j}}{\sum_{k=1}^{N} \exp(A_{i,k}) M_{i,k}}, 
    \\ \quad 1 \leq i, j \leq N,
\end{multline}
where $(\boldsymbol{l+x}) \in \mathbb{N} < l'$. $\boldsymbol{l'}$ is the position layer of the next decision module. If $M^L_{i,j} = 0$, the attention score for token $H_j$ will be $0$ in the final attention matrix, resulting in $H_j$ won't contribute to any other tokens. In addition, we define $\boldsymbol{D_l}, \boldsymbol{D_{l'}}$ as the token pruning decisions get from layer $l$, $l'$, respectively and $l' > l$. We update $\boldsymbol{D_{l'}}$ by 
\begin{equation}
\boldsymbol{D_{l'}} \leftarrow D_{l} \odot D_{l'},
\end{equation}
where $\odot$ is element-wise production, which means a previously removed vision token will never be used again.

In conclusion, Equations \ref{equ8} - \ref{equ13} remove the effects of unwanted vision tokens on other tokens while keeping the number of total tokens unchanged. By Equations \ref{equ6},\ref{equ7},\ref{equ12},\ref{equ13}, the process of generating and applying token pruning decisions is fully differentiable. These two factors achieve the end-to-end training capability of LVPruning.

\textbf{Training Objectives}:
The training objectives of LVPruning are designed to teach the decision modules to remove vision tokens to predetermined ratios at different layers while fine-tuning the MLLMs to maintain their vision instruction-following capability despite the token pruning. The primary training objective is causal language modeling for instruction tuning, as described by \citet{DBLP:conf/nips/VaswaniSPUJGKP17, DBLP:conf/nips/LiuLWL23a, DBLP:journals/corr/abs-2307-09288}. Since causal language modeling is a widely used loss function, we do not detail its formal definition in this paper, referring to it as $\boldsymbol{\mathcal{L}_{causal}}$.

Additionally, to ensure that the ratio of retained vision tokens aligns with predefined values at each decision module, we insert $\boldsymbol{S}$ decision modules into the LLM at specific layer indices $\boldsymbol{L_{idx}}=[l_1,...,l_S]$, with target token retention ratios $\boldsymbol{\mathrm{P}} = [\rho_1, ... , \rho_S]$. To enforce this, we apply Mean Squared Error (MSE) loss $\mathcal{L}_{\text{ratio}}$ to constrain the token pruning decisions:
\begin{equation}
\mathcal{L}_{\text{ratio}} = \frac{1}{S} \sum_{s=1}^{S} ( \rho_s - \frac{1}{|I_V|} \sum_{i=1}^{|I_V|} D_{l_s, i})^2,
\end{equation}
where \( \delta(D_{l_s}, \rho_s) \) is the Huber loss, $\beta$ is a threshold that determines the loss function used. We set $\beta=0.5$ in all our experiments. The final training objective is the weighted sum of $\boldsymbol{\mathcal{L}_{causal}}$ and $\boldsymbol{\mathcal{L}}_{\text{ratio}}$:
\begin{equation}
    \boldsymbol{\mathcal{L}} = \lambda_{causal}\mathcal{L}_{causal} + \lambda_{ratio}\mathcal{L}_{ratio}.
\end{equation}

\subsection{Inference}

During the training phase, attention masks are employed to exclude the impact of irrelevant vision tokens. However, during inference, it is necessary to remove these tokens to reduce computational expenses, which introduces significant practical difficulties. First, the quantity of retained vision tokens, as determined by $\boldsymbol{D}$, is variable, thereby complicating the process of batch inference. Second, contemporary LLMs generally utilize positional embeddings for tokens at each layer. It is essential to maintain the original positional embeddings for the retained tokens to ensure alignment with the distribution seen during training.

To overcome the first issue, we define a set of token kept ratios $\hat{\boldsymbol{\mathrm{P}}} = [\hat{\rho_1}, ... ,\hat{\rho_S}]$ during inference. Note that the inference ratios do not have to be the same as the training ratios. At the $s$-th token pruning layer, we first sort the decision scores
\begin{equation}
    \boldsymbol{\mathrm{Q}^s} = argsort(D_{ls}).
\end{equation}
Then keep the top $\boldsymbol{k_s} = \rho_s \times \mid I_V \mid $ vision tokens with the highest scores. The kept vision token indices among all vision tokens are $\boldsymbol{\hat{I}^s}= \{Q^s_{1:k_s}\}$, and the kept vision token indices among all tokens are $\boldsymbol{\hat{I}^{s}_v}=I_{v,{\hat{I}^s}}$. We define $\boldsymbol{\mathrm{PE}^1}$ as the positional embeddings at the first layer. To ensure that the positional embeddings for both vision and text tokens remain unchanged after each pruning, the positional embeddings at the $s$-th token pruning layer are
\begin{equation}
    \boldsymbol{PE^s} = [PE^1_{\hat{I}^s_v}, PE^1_{I_T}].
\end{equation}

\section{Experimental Setup}

The objective of our experiments is to investigate the feasibility of employing token pruning techniques to enhance the efficiency of MLLMs. Specifically, we ask:  \emph{(i)} Does LVPruning effectively reduce the computational costs while keeping the performance unchanged and to what extent can vision tokens be pruned? \emph{(ii)} Compared with state-of-the-art MLLMs, does LVPruning achieve a balance between computational costs and model performance? To answer question \emph{(i)} we compare LVPruning with the base MLLM on various benchmarks using different vision token kept ratios. To answer \emph{ii}, we compare the relationship between inference TFLOPs and model performance for LVPruning and various state-of-the-art MLLMs.


\subsection{Implementation Details}

In all our experiments, we apply LVPruning to LLaVA-1.5-7B \citep{DBLP:journals/corr/abs-2310-03744} (hereafter referred as LVPruning) by inserting $S=3$ decision modules with token kept ratio $ \boldsymbol{\mathrm{P}} = [\rho , \rho-0.2, \rho-0.4]$, where $\rho=0.5$ for training. These modules are inserted after the 1st, 8th, and 16th layers of LLaMA LLM. In each token pruning layer, we utilize 2 sequential cross-attention blocks, each with 8 attention heads. The Feed-Forward Networks (FFNs) in these cross-attention blocks follow the architecture: [LayerNorm, Linear(C, 2C), SiLu activation, Linear(2C, C), LayerNorm]. We follow most of LLaVA-1.5’s training settings, freezing all parameters from the LLaVA and only training inserted modules. The learning rate is set to 2e-6, with a 0.03 warm-up ratio and a cosine learning rate scheduler. The batch size is 64, and no weight decay is applied. Additionally, a maximum gradient norm of 1.0 is used to stabilize convergence. Training runs on 8 A100 (80G) GPUs. During inference, we evaluate using three different token kept ratios ($\rho=0.6$, $\rho=0.5$, and $\rho=0.45$) without tuning any model parameters. All inference TFLOPs reported in this paper are computed using a dummy input consisting of 1 image and 30 text tokens.

\subsection{Dataset and Benchmarks}
To prove LVPruning's data efficiency, we use a subset of the training data for LLaVA-1.5. The LLaVA-1.5 Vision Instruction Tuning dataset \citep{DBLP:journals/corr/abs-2310-03744} consists of 665k data samples. We remove all entries without image inputs, which results in approximately 620k training samples, and the model is trained for 1 epoch. To evaluate the performance and computational efficiency of LVPruning, we calculate its inference FLOPs and assess it on nine multi-modal benchmark datasets. These include VQAv2 \citep{DBLP:conf/cvpr/GoyalKSBP17}, GQA \citep{DBLP:conf/cvpr/HudsonM19}, VizWiz \citep{DBLP:conf/cvpr/Gurari0SGLGLB18}, SciQA-IMG \citep{DBLP:conf/nips/LuMX0CZTCK22}, and TextQA \citep{DBLP:conf/cvpr/SinghNSJCBPR19}, with top1 accuracy (acc@1) used as the evaluation metric for all these benchmarks. Additionally, POPE \citep{DBLP:conf/emnlp/LiDZWZW23} is assessed using the F1 score across three splits. MMBench \citep{DBLP:journals/corr/abs-2307-06281} is evaluated through multiple-choice questions on both English and Chinese-translated versions. LLaVA-Wild \citep{DBLP:conf/nips/LiuLWL23a} and MM-Vet \citep{DBLP:conf/icml/YuYLWL0WW24} assess model responses with the assistance of GPT-4. The details about each benchmark can be found in Appendix \ref{app:benchmarks}.

\begin{table*}[t]
    \centering
    \begin{tabular}{@{}l|c|ccccc@{}}
        \toprule
        Method & TFLOPs & VQAv2 & GQA & Vizwiz & SQA-IMG & TextVQA \\ \midrule
        LLaVA-1.5-7B & 8.38 & 78.5 & 62.0 & 50.0 & 69.4* & 58.2 \\\midrule
        LVPruning ($\rho=0.6$) & 
        ${3.97}_{\textcolor{red}{-52.6\%}}$ & 
        ${78.1}_{\textcolor{red}{-0.4}}$ & 
        ${61.5}_{\textcolor{red}{-0.5}}$ & 
        ${51.2}_{\textcolor{blue}{+1.2}}$ & 
        ${69.0}_{\textcolor{red}{-0.4}}$ & 
        ${58.2}_{\textcolor{blue}{+0}}$ \\
        LVPruning ($\rho=0.5$)& 
        ${3.18}_{\textcolor{red}{-62.1\%}}$ & 
        $77.3_{\textcolor{red}{-1.2}}$ & 
        $60.7_{\textcolor{red}{-1.3}}$ & 
        $51.3_{\textcolor{blue}{+1.3}}$ & 
        $68.7_{\textcolor{red}{-0.7}}$ & 
        $58.0_{\textcolor{red}{-0.2}}$ \\
        LVPruning ($\rho=0.45$)& 
        ${2.79}_{\textcolor{red}{-66.7\%}} $& 
        $75.7_{\textcolor{red}{-2.8}}$ & 
        $59.3_{\textcolor{red}{-2.7}}$ & 
        $50.8_{\textcolor{blue}{+0.8}}$ & 
        $68.6_{\textcolor{red}{-0.8}}$ & 
        $57.5_{\textcolor{red}{-0.7}}$ \\ 
        \bottomrule
    \end{tabular}
    \caption{Performance and inference TFLOPs comparison between LLaVA-1.5-7B \citep{DBLP:journals/corr/abs-2310-03744} and LVPruning on Visual Question Answering (VQA) benchmarks. LVPruning can significantly reduce the inference cost while maintaining marginal performance loss.}
    \label{tab:main_res1}

    \vspace{0.2cm}
    \centering

    \begin{tabular}{@{}l|ccc|cc|c|c@{}}
        \toprule
        \multirow{2}{*}{Method} & 
        \multicolumn{3}{c|}{POPE} & 
        \multicolumn{2}{c|}{MMBench} & 
        LLaVA & 
        MM \\ 
        & rad & pop & adv & en & cn  & -Wild & -Vet \\ 
        \midrule
        LLaVA-1.5-7B& 
        87.3 & 
        86.1 & 
        84.2 & 
        64.3& 
        58.3& 
        65.4& 
        31.1 \\
        \midrule
        LVPruning ($\rho=0.6$)   & 
        $88.1_{\textcolor{blue}{+0.8}}$ & 
        $86.5_{\textcolor{blue}{+0.4}}$ & 
        $84.4_{\textcolor{blue}{+0.2}}$ & 
        $64.0_{\textcolor{red}{-0.3}}$ & 
        $57.4_{\textcolor{red}{-0.9}}$ & 
        $67.9_{\textcolor{blue}{+2.5}}$ & 
        $33.3_{\textcolor{blue}{+2.2}}$ \\
        LVPruning ($\rho=0.5$) & 
        $87.6_{\textcolor{blue}{+0.3}}$ & 
        $86.2_{\textcolor{blue}{+0.1}}$ & 
        $84.1_{\textcolor{red}{-0.1}}$ & 
        $63.9_{\textcolor{red}{-0.4}}$ & 
        $56.3_{\textcolor{red}{-2.0}}$ &
        $65.5_{\textcolor{blue}{+0.1}}$ & 
        $31.6_{\textcolor{blue}{+0.5}}$ \\
        LVPruning ($\rho=0.45$) & 
        $87.4_{\textcolor{blue}{+0.1}}$ & 
        $86.1_{\textcolor{blue}{+0}}$ & 
        $84.0_{\textcolor{red}{-0.2}}$ & 
        $63.4_{\textcolor{red}{-0.9}}$ & 
        $56.3_{\textcolor{red}{-2.0}}$ & 
        $60.7_{\textcolor{red}{-4.7}}$ & 
        $30.8_{\textcolor{red}{-0.3}}$ \\
        \bottomrule
    \end{tabular}
    \caption{Performance comparison between LLaVA-1.5-7B \citep{DBLP:journals/corr/abs-2310-03744} and LVPruning on visual instruction following benchmarks. LVPruning achieves competitive results, with improvements observed on three benchmarks.}
    \label{tab:main_res2}
    \vspace{0.2cm}
\end{table*}

\begin{figure}[t]
\centering
    \includegraphics[scale=0.38]{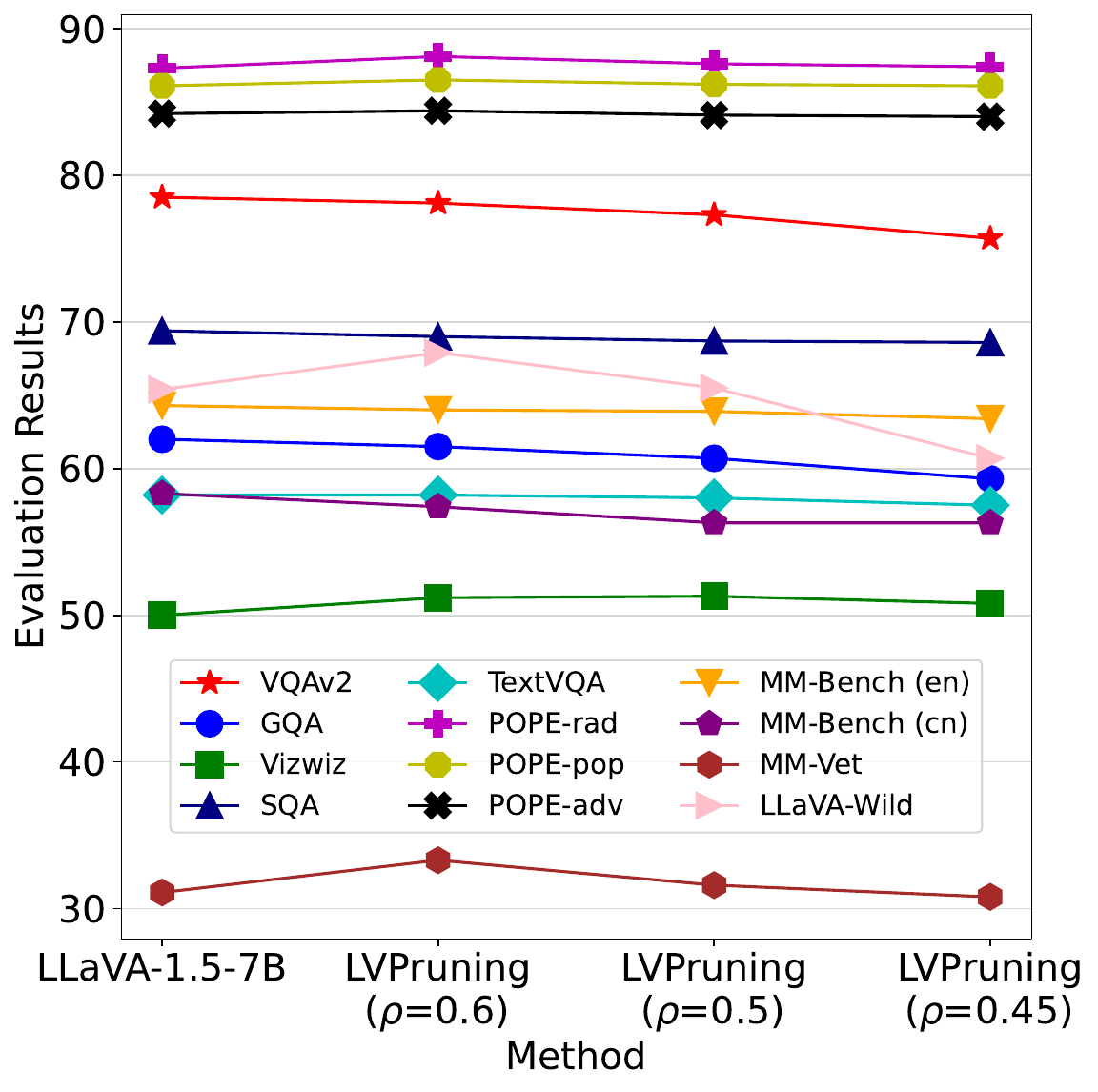}
    \caption{Performance variance between LLaVA-1.5-7B \citep{DBLP:journals/corr/abs-2310-03744} and LVPruning with different vision token kept ratios $\rho$ on nine multi-modal benchmarks. Even at a low token kept ratio, such as $\rho=0.45$, the performance degradation remains small.}
    \label{fig:7b-different_ratio}
\end{figure}

\begin{figure}[t]
\centering
    \includegraphics[scale=0.38]{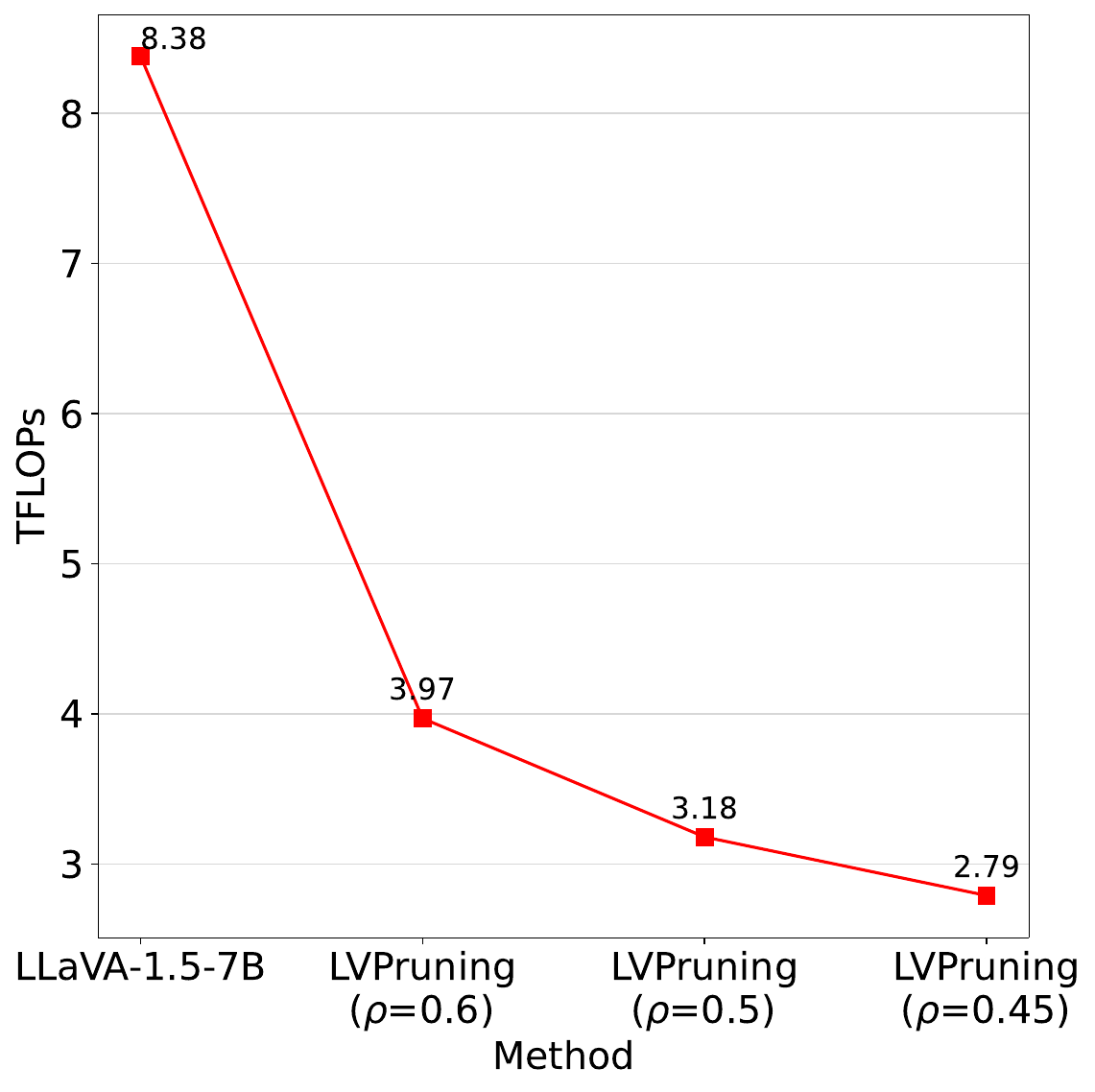}
    \caption{Comparison of Inference FLOPs of LLaVA-1.5-7B \citep{DBLP:journals/corr/abs-2310-03744} and LVPruning with different vision token kept ratio $\rho$.}
    \label{fig:7b-different_tflops}
\end{figure}

\begin{table*}[t]
 \centering
    \begin{tabular}{@{}lc|ccccc@{}}
        \toprule
        Method & TFLOPs & VQAv2 & GQA & Vizwiz & SQA-IMG & TextVQA \\
        \midrule
        BLIP2-14B & 2.14 & 65.0 &41.0 & 19.6 & 61 & 42.5 \\
        InstructBLIP-8B &  1.36 & – & 49.2 & 34.5 & 60.5 & 50.1 \\
        InstructBLIP-14B &  2.14 & – & 49.5 & 33.4 & 63.1 & 50.7 \\
        IDEFICS-9B &  0.87 & 50.9 & 38.4 & 35.5 & – & 25.9 \\
        Qwen-VL &  1.75 & \textbf{78.8} & 59.3 & 35.2 & 67.1 & \textbf{63.8} \\
        Qwen-VL-Chat &  1.75 & 78.2 & 57.5 & 38.9 & 68.2 & 61.5 \\
        \midrule
        LVPruning ($\rho=0.5$) &  3.18 & 77.3 & \textbf{60.7} & \textbf{51.34} & \textbf{68.7} & 58.0\\
        \bottomrule
    \end{tabular}
    \caption{Performance comparison between LVPruning with a token retention ratio of $\rho=0.5$ and state-of-the-art Q-former-based models on Visual Question Answering (VQA) benchmarks. It shows that LVPruning ($\rho=0.5$) outperforms the state-of-the-art MLLMs on three benchmarks, while delivering competitive results on the others.}
    \label{tab:3}
    \vspace{0.2cm}

    \centering

    \begin{tabular}{@{}l|ccc|cc|c|c@{}}
        \toprule
        \multirow{2}{*}{Method} & \multicolumn{3}{c|}{POPE} & \multicolumn{2}{c|}{MMBench} & LLaVA & MM \\ 
        & rad & pop & adv & en & cn  & -Wild & -Vet \\ 
        \midrule
        BLIP2-14B      & \textbf{89.6}    & 85.5    & 80.9    & - & - & 38.1 & 22.4 \\
        InstructBLIP-8B      & -    & -    & -    & 36.0 & 23.7 & 60.9 & 26.2 \\
        InstructBLIP-14B      & 87.7    & 77    & 72    & - & - & 58.2 & 25.6 \\
        IDEFICS-9B           & -    & -    & -    & 48.2 & 25.2 & -    & -    \\
        Qwen-VL              & -    & -    & -    & 38.2 & 7.4  & -    & -    \\
        Qwen-VL-Chat         & -    & -    & -    & 60.6 & 56.7 & -    & -    \\
        \midrule
        LVPruning ($\rho=0.5$)   & 87.6 & \textbf{86.2} & \textbf{84.1} & \textbf{63.9} & \textbf{56.3} & \textbf{65.5} & \textbf{31.6} \\
        \bottomrule
    \end{tabular}
    \caption{Performance comparison between LVPruning with a token retention ratio of $\rho=0.5$ and state-of-the-art Q-former-based models on visual instruction following benchmarks. It shows that LVPruning outperforms nearly all benchmarks, with only a slight performance drop compared to BLIP2-14B \citep{DBLP:conf/icml/0008LSH23} in POPE(rad)\citep{DBLP:conf/emnlp/LiDZWZW23}.}
    \label{tab:4}
    \vspace{0.2cm}
\end{table*}

\section{Experimental Results}

In this section, we analyze the experimental results of LVPruning across nine multi-modal benchmarks. Section \ref{sec:res1} examines the performance and inference TFLOPs of LVPruning compared to the base MLLM, LLaVA-1.5 \citep{DBLP:journals/corr/abs-2310-03744}, highlighting its effectiveness in reducing computational cost. Section \ref{sec:res2} compares LVPruning with state-of-the-art Q-former-based MLLMs, demonstrating its ability to balance performance and efficiency.

\subsection{Performance Preservation and Computation Cost Reduction}\label{sec:res1}

To address research question \emph{(i)}, we compare the model performance and inference TFLOPs of LVPruning at various vision token kept ratios, evaluating both computational savings and performance trade-off. Generally speaking, Table \ref{tab:main_res1} and Table \ref{tab:main_res2} show that LVPruning significantly reduces the inference cost while maintaining competitive performance. As shown in Table \ref{tab:main_res1}, on Visual Question Answering (VQA) benchmarks, with a pruning ratio of $\rho=0.6$, LVPruning achieves a 52.6\% reduction in TFLOPs, dropping from 8.38 to 3.97 TFLOPs, with only a minor performance degradation. For instance, on VQAv2, GQA, and VizWiz benchmarks, LVPruning maintains similar accuracy, with minimal decreases of 0.4, 0.5, and even an increase of 1.2 points, respectively. Even with higher pruning ratios, such as $\rho=0.45$ (66.7\% TFLOPs reduction), the model's performances on certain benchmarks like VizWiz (+0.8) and SQA-IMG (-0.8) are still comparable to LLaVA-1.5-7B. Table \ref{tab:main_res2} shows the results for visual instruction following benchmarks. On the POPE and MMBench (en and cn) benchmarks, LVPruning yields similar or improved performance with performance gains of up to 0.8 points on POPE. Notably, the LLaVA-Wild and MM-Vet benchmarks show obvious performance gains with $\rho=0.6$ which increases by 2.5 and 2.2 points, respectively.

Overall, Figure \ref{fig:7b-different_ratio} visualizes the performance variance between LLaVA-1.5 \citep{DBLP:journals/corr/abs-2310-03744} and LVPruning with different vision token kept ratios $\rho$ on nine multi-modal benchmarks, and Figure \ref{fig:7b-different_tflops} demonstrates their inference FLOPs. LVPruning demonstrates that significant computational savings can be achieved with minimal impact on performance. Even at a high pruning ratio, such as $\rho=0.45$, where TFLOPs are reduced by 66.7\%, the performance degradation remains relatively small, indicating that LVPruning effectively balances model efficiency with task performance. More details about the relationship between token kept ratios and inference FLOPs are shown in Appendix \ref{app:ratio_flops}.

\begin{figure}[t]
\centering
        \includegraphics[scale=0.46]{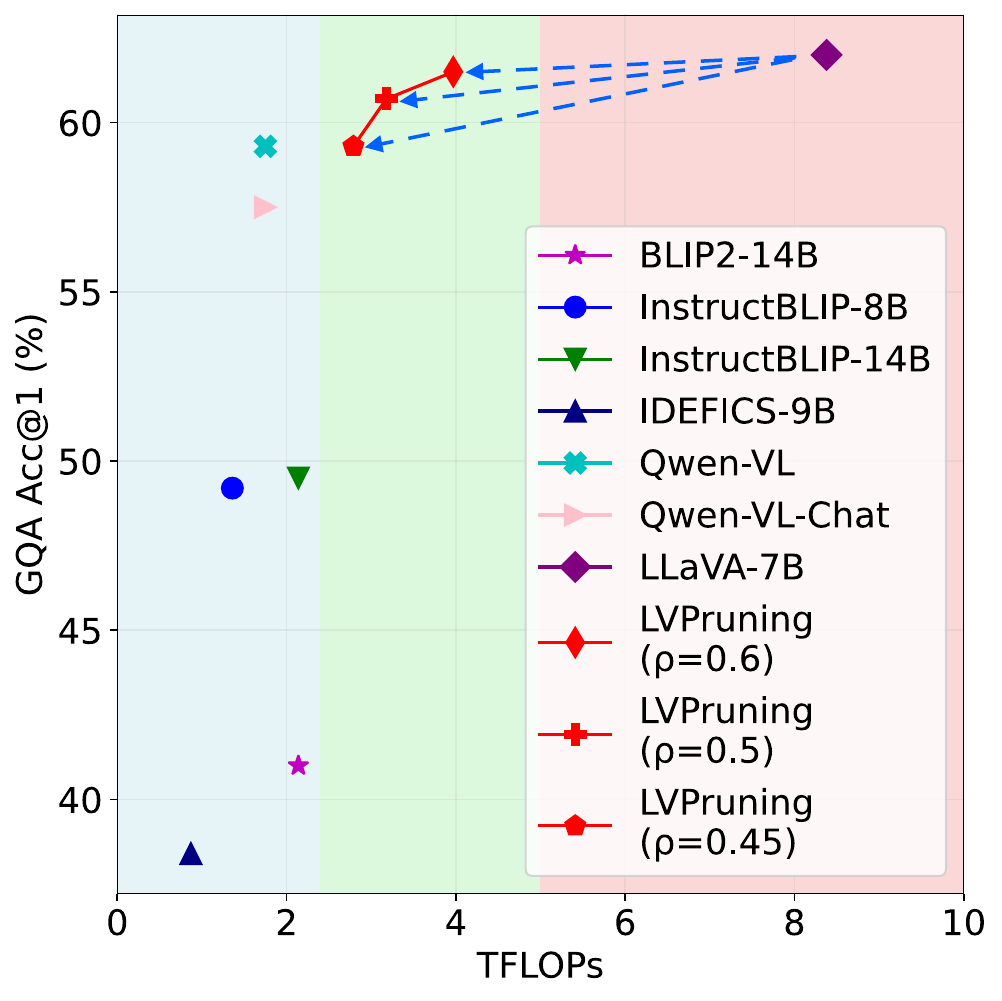}
        \caption{The relationship between inference TFLOPs and the performance of LVPruning and other state-of-the-art MLLMs evaluated on the GQA benchmark \citep{DBLP:conf/cvpr/HudsonM19}. LVPruning (green area) outperforms Q-former-based models (blue area) while achieving substantial computational savings compared to LLaVA-1.5 (red area) \citep{DBLP:journals/corr/abs-2310-03744}. This demonstrates LVPruning’s ability to balance performance and efficiency among state-of-the-art MLLMs.}
    \label{fig:sota_flops}
\end{figure}

\subsection{Comparisons with state-of-the-art MLLMs} \label{sec:res2}
To address research question \emph{(ii)}, we use LVPruning with $\rho=0.5$ as a representative configuration to compare against state-of-the-art Q-former-based MLLMs. Generally speaking, Tables \ref{tab:3} and \ref{tab:4} demonstrate that LVPruning achieves superior performance with competitive inference FLOPs.

As shown in Table \ref{tab:3}, on VQA benchmarks, LVPruning with $\rho=0.5$ outperforms several state-of-the-art models, such as BLIP2-14B \citep{DBLP:conf/icml/0008LSH23}, InstructBLIP-14B \citep{DBLP:conf/nips/Dai0LTZW0FH23}, and IDEFICS-9B \citep{DBLP:conf/nips/LaurenconSTBSLW23}. For example, LVPruning achieves a higher VQAv2 accuracy (77.3) compared to BLIP2-14B (65.0) and IDEFICS-9B (50.9), while utilizing only 3.18 TFLOPs, which is relatively higher than IDEFICS but remains efficient compared to other models like BLIP2 and Qwen-VL \citep{Bai_Bai_Yang_Wang_Tan_Wang_Lin_Zhou_Zhou_2023a}. In tasks such as VizWiz, LVPruning also achieves a notable boost in performance (51.34), surpassing BLIP2 and InstructBLIP models by a large margin. Table \ref{tab:4} shows that on visual instruction following benchmarks, LVPruning consistently delivers competitive results. For instance, LVPruning achieves 87.6 accuracy on the POPE (rad) benchmark, closely trailing BLIP2-14B's 89.6. However, it outperforms BLIP2 on multiple datasets, such as MMBench (en) and LLaVA-Wild, with scores of 63.9 and 65.5, respectively. These results illustrate that LVPruning maintains competitive performance across various instruction following benchmarks, balancing between efficiency and effectiveness. Figure \ref{fig:sota_flops} further illustrates the relationship between inference TFLOPs and the performance of LVPruning and state-of-the-art MLLMs on the GQA benchmark. LVPruning, represented by red points, showcases a balanced trade-off between computational efficiency and performance.

\section{Conclusion}
In this work, we introduce LVPruning, a novel language-guided vision token pruning method that can be integrated into existing MLLMs with minimal architectural changes. LVPruning computes relevance scores for each vision token based on language tokens, progressively removing redundant tokens throughout the LLM. By the middle layer, it eliminates up to 90\% of vision tokens, achieving a 62.1\% reduction in FLOPs with only a $\sim$0.45\% average performance loss across nine multi-modal benchmarks. This makes LVPruning a practical solution for enhancing MLLM efficiency while preserving performance in multi-modal tasks.

\section{Limitations}

While LVPruning shows significant promise in reducing computational load, several limitations to our study should be acknowledged. Our evaluation has been conducted on a specific set of benchmarks. The performance of LVPruning on other datasets or in real-world applications remains unexplored. Therefore, although LVPruning is effective in reducing computational overhead, it is crucial to consider the specific requirements of different tasks when applying this method to ensure that essential visual information is not compromised. Future research could involve evaluating the performance of LVPruning with human feedback to better understand its practical implications. 

\bibliography{custom.bib}

\begin{thebibliography}{29}
\providecommand{\natexlab}[1]{#1}

\bibitem[{Bai et~al.(2023)Bai, Bai, Yang, Wang, Tan, Wang, Lin, Zhou, and Zhou}]{Bai_Bai_Yang_Wang_Tan_Wang_Lin_Zhou_Zhou_2023a}
Jinze Bai, Shuai Bai, Shusheng Yang, Shijie Wang, Sinan Tan, Peng Wang, Junyang Lin, Chang Zhou, and Jingren Zhou. 2023.
\newblock \href {https://openreview.net/forum?id=qrGjFJVl3m} {Qwen-vl: A versatile vision-language model for understanding...}

\bibitem[{Bolya et~al.(2023)Bolya, Fu, Dai, Zhang, Feichtenhofer, and Hoffman}]{DBLP:conf/iclr/BolyaFDZFH23}
Daniel Bolya, Cheng{-}Yang Fu, Xiaoliang Dai, Peizhao Zhang, Christoph Feichtenhofer, and Judy Hoffman. 2023.
\newblock Token merging: Your vit but faster.
\newblock In \emph{{ICLR}}. OpenReview.net.

\bibitem[{Chen et~al.(2023)Chen, Shao, Xu, Lin, Zhang, Chao, Ji, Qiao, and Luo}]{DBLP:conf/iccv/ChenSXLZCJQL23}
Mengzhao Chen, Wenqi Shao, Peng Xu, Mingbao Lin, Kaipeng Zhang, Fei Chao, Rongrong Ji, Yu~Qiao, and Ping Luo. 2023.
\newblock Diffrate : Differentiable compression rate for efficient vision transformers.
\newblock In \emph{{ICCV}}, pages 17118--17128. {IEEE}.

\bibitem[{Dai et~al.(2023)Dai, Li, Li, Tiong, Zhao, Wang, Li, Fung, and Hoi}]{DBLP:conf/nips/Dai0LTZW0FH23}
Wenliang Dai, Junnan Li, Dongxu Li, Anthony Meng~Huat Tiong, Junqi Zhao, Weisheng Wang, Boyang Li, Pascale Fung, and Steven C.~H. Hoi. 2023.
\newblock Instructblip: Towards general-purpose vision-language models with instruction tuning.
\newblock In \emph{NeurIPS}.

\bibitem[{Gong et~al.(2014)Gong, Liu, Yang, and Bourdev}]{DBLP:journals/corr/GongLYB14}
Yunchao Gong, Liu Liu, Ming Yang, and Lubomir~D. Bourdev. 2014.
\newblock Compressing deep convolutional networks using vector quantization.
\newblock \emph{CoRR}, abs/1412.6115.

\bibitem[{Goyal et~al.(2017)Goyal, Khot, Summers{-}Stay, Batra, and Parikh}]{DBLP:conf/cvpr/GoyalKSBP17}
Yash Goyal, Tejas Khot, Douglas Summers{-}Stay, Dhruv Batra, and Devi Parikh. 2017.
\newblock Making the {V} in {VQA} matter: Elevating the role of image understanding in visual question answering.
\newblock In \emph{{CVPR}}, pages 6325--6334. {IEEE} Computer Society.

\bibitem[{Gurari et~al.(2018)Gurari, Li, Stangl, Guo, Lin, Grauman, Luo, and Bigham}]{DBLP:conf/cvpr/Gurari0SGLGLB18}
Danna Gurari, Qing Li, Abigale~J. Stangl, Anhong Guo, Chi Lin, Kristen Grauman, Jiebo Luo, and Jeffrey~P. Bigham. 2018.
\newblock Vizwiz grand challenge: Answering visual questions from blind people.
\newblock In \emph{{CVPR}}, pages 3608--3617. Computer Vision Foundation / {IEEE} Computer Society.

\bibitem[{Hinton et~al.(2015)Hinton, Vinyals, and Dean}]{DBLP:journals/corr/HintonVD15}
Geoffrey~E. Hinton, Oriol Vinyals, and Jeffrey Dean. 2015.
\newblock Distilling the knowledge in a neural network.
\newblock \emph{CoRR}, abs/1503.02531.

\bibitem[{Hudson and Manning(2019)}]{DBLP:conf/cvpr/HudsonM19}
Drew~A. Hudson and Christopher~D. Manning. 2019.
\newblock {GQA:} {A} new dataset for real-world visual reasoning and compositional question answering.
\newblock In \emph{{CVPR}}, pages 6700--6709. Computer Vision Foundation / {IEEE}.

\bibitem[{Lauren{\c{c}}on et~al.(2023)Lauren{\c{c}}on, Saulnier, Tronchon, Bekman, Singh, Lozhkov, Wang, Karamcheti, Rush, Kiela, Cord, and Sanh}]{DBLP:conf/nips/LaurenconSTBSLW23}
Hugo Lauren{\c{c}}on, Lucile Saulnier, L{\'{e}}o Tronchon, Stas Bekman, Amanpreet Singh, Anton Lozhkov, Thomas Wang, Siddharth Karamcheti, Alexander~M. Rush, Douwe Kiela, Matthieu Cord, and Victor Sanh. 2023.
\newblock {OBELICS:} an open web-scale filtered dataset of interleaved image-text documents.
\newblock In \emph{NeurIPS}.

\bibitem[{Li et~al.(2023{\natexlab{a}})Li, Li, Savarese, and Hoi}]{DBLP:conf/icml/0008LSH23}
Junnan Li, Dongxu Li, Silvio Savarese, and Steven C.~H. Hoi. 2023{\natexlab{a}}.
\newblock {BLIP-2:} bootstrapping language-image pre-training with frozen image encoders and large language models.
\newblock In \emph{{ICML}}, volume 202 of \emph{Proceedings of Machine Learning Research}, pages 19730--19742. {PMLR}.

\bibitem[{Li et~al.(2023{\natexlab{b}})Li, Du, Zhou, Wang, Zhao, and Wen}]{DBLP:conf/emnlp/LiDZWZW23}
Yifan Li, Yifan Du, Kun Zhou, Jinpeng Wang, Wayne~Xin Zhao, and Ji{-}Rong Wen. 2023{\natexlab{b}}.
\newblock Evaluating object hallucination in large vision-language models.
\newblock In \emph{{EMNLP}}, pages 292--305. Association for Computational Linguistics.

\bibitem[{Liang et~al.(2022)Liang, Ge, Tong, Song, Wang, and Xie}]{DBLP:journals/corr/abs-2202-07800}
Youwei Liang, Chongjian Ge, Zhan Tong, Yibing Song, Jue Wang, and Pengtao Xie. 2022.
\newblock Not all patches are what you need: Expediting vision transformers via token reorganizations.
\newblock \emph{CoRR}, abs/2202.07800.

\bibitem[{Liu et~al.(2023{\natexlab{a}})Liu, Li, Li, and Lee}]{DBLP:journals/corr/abs-2310-03744}
Haotian Liu, Chunyuan Li, Yuheng Li, and Yong~Jae Lee. 2023{\natexlab{a}}.
\newblock Improved baselines with visual instruction tuning.
\newblock \emph{CoRR}, abs/2310.03744.

\bibitem[{Liu et~al.(2023{\natexlab{b}})Liu, Li, Wu, and Lee}]{DBLP:conf/nips/LiuLWL23a}
Haotian Liu, Chunyuan Li, Qingyang Wu, and Yong~Jae Lee. 2023{\natexlab{b}}.
\newblock \href {http://papers.nips.cc/paper\_files/paper/2023/hash/6dcf277ea32ce3288914faf369fe6de0-Abstract-Conference.html} {Visual instruction tuning}.
\newblock In \emph{Advances in Neural Information Processing Systems 36: Annual Conference on Neural Information Processing Systems 2023, NeurIPS 2023, New Orleans, LA, USA, December 10 - 16, 2023}.

\bibitem[{Liu et~al.(2023{\natexlab{c}})Liu, Duan, Zhang, Li, Zhang, Zhao, Yuan, Wang, He, Liu, Chen, and Lin}]{DBLP:journals/corr/abs-2307-06281}
Yuan Liu, Haodong Duan, Yuanhan Zhang, Bo~Li, Songyang Zhang, Wangbo Zhao, Yike Yuan, Jiaqi Wang, Conghui He, Ziwei Liu, Kai Chen, and Dahua Lin. 2023{\natexlab{c}}.
\newblock Mmbench: Is your multi-modal model an all-around player?
\newblock \emph{CoRR}, abs/2307.06281.

\bibitem[{Lu et~al.(2022)Lu, Mishra, Xia, Qiu, Chang, Zhu, Tafjord, Clark, and Kalyan}]{DBLP:conf/nips/LuMX0CZTCK22}
Pan Lu, Swaroop Mishra, Tanglin Xia, Liang Qiu, Kai{-}Wei Chang, Song{-}Chun Zhu, Oyvind Tafjord, Peter Clark, and Ashwin Kalyan. 2022.
\newblock Learn to explain: Multimodal reasoning via thought chains for science question answering.
\newblock In \emph{NeurIPS}.

\bibitem[{Radford et~al.(2021)Radford, Kim, Hallacy, Ramesh, Goh, Agarwal, Sastry, Askell, Mishkin, Clark, Krueger, and Sutskever}]{DBLP:conf/icml/RadfordKHRGASAM21}
Alec Radford, Jong~Wook Kim, Chris Hallacy, Aditya Ramesh, Gabriel Goh, Sandhini Agarwal, Girish Sastry, Amanda Askell, Pamela Mishkin, Jack Clark, Gretchen Krueger, and Ilya Sutskever. 2021.
\newblock \href {http://proceedings.mlr.press/v139/radford21a.html} {Learning transferable visual models from natural language supervision}.
\newblock In \emph{Proceedings of the 38th International Conference on Machine Learning, {ICML} 2021, 18-24 July 2021, Virtual Event}, volume 139 of \emph{Proceedings of Machine Learning Research}, pages 8748--8763. {PMLR}.

\bibitem[{Rao et~al.(2021)Rao, Zhao, Liu, Lu, Zhou, and Hsieh}]{DBLP:conf/nips/RaoZLLZH21}
Yongming Rao, Wenliang Zhao, Benlin Liu, Jiwen Lu, Jie Zhou, and Cho{-}Jui Hsieh. 2021.
\newblock Dynamicvit: Efficient vision transformers with dynamic token sparsification.
\newblock In \emph{NeurIPS}, pages 13937--13949.

\bibitem[{Sanh et~al.(2019)Sanh, Debut, Chaumond, and Wolf}]{DBLP:journals/corr/abs-1910-01108}
Victor Sanh, Lysandre Debut, Julien Chaumond, and Thomas Wolf. 2019.
\newblock Distilbert, a distilled version of {BERT:} smaller, faster, cheaper and lighter.
\newblock \emph{CoRR}, abs/1910.01108.

\bibitem[{Singh et~al.(2019)Singh, Natarajan, Shah, Jiang, Chen, Batra, Parikh, and Rohrbach}]{DBLP:conf/cvpr/SinghNSJCBPR19}
Amanpreet Singh, Vivek Natarajan, Meet Shah, Yu~Jiang, Xinlei Chen, Dhruv Batra, Devi Parikh, and Marcus Rohrbach. 2019.
\newblock Towards {VQA} models that can read.
\newblock In \emph{{CVPR}}, pages 8317--8326. Computer Vision Foundation / {IEEE}.

\bibitem[{Sun et~al.(2024)Sun, Li, Lin, and Batista{-}Navarro}]{DBLP:conf/haii/SunLLB24}
Yizheng Sun, Hao Li, Chenghua Lin, and Riza Batista{-}Navarro. 2024.
\newblock Lanvikd: Cross-modal language-vision knowledge distillation for egocentric action recognition.
\newblock In \emph{HAII5.0@ECAI}, volume 3765 of \emph{{CEUR} Workshop Proceedings}. CEUR-WS.org.

\bibitem[{Touvron et~al.(2023)Touvron, Martin, Stone, Albert, Almahairi, Babaei, Bashlykov, Batra, Bhargava, Bhosale, Bikel, Blecher, Canton{-}Ferrer, Chen, Cucurull, Esiobu, Fernandes, Fu, Fu, Fuller, Gao, Goswami, Goyal, Hartshorn, Hosseini, Hou, Inan, Kardas, Kerkez, Khabsa, Kloumann, Korenev, Koura, Lachaux, Lavril, Lee, Liskovich, Lu, Mao, Martinet, Mihaylov, Mishra, Molybog, Nie, Poulton, Reizenstein, Rungta, Saladi, Schelten, Silva, Smith, Subramanian, Tan, Tang, Taylor, Williams, Kuan, Xu, Yan, Zarov, Zhang, Fan, Kambadur, Narang, Rodriguez, Stojnic, Edunov, and Scialom}]{DBLP:journals/corr/abs-2307-09288}
Hugo Touvron, Louis Martin, Kevin Stone, Peter Albert, Amjad Almahairi, Yasmine Babaei, Nikolay Bashlykov, Soumya Batra, Prajjwal Bhargava, Shruti Bhosale, Dan Bikel, Lukas Blecher, Cristian Canton{-}Ferrer, Moya Chen, Guillem Cucurull, David Esiobu, Jude Fernandes, Jeremy Fu, Wenyin Fu, Brian Fuller, Cynthia Gao, Vedanuj Goswami, Naman Goyal, Anthony Hartshorn, Saghar Hosseini, Rui Hou, Hakan Inan, Marcin Kardas, Viktor Kerkez, Madian Khabsa, Isabel Kloumann, Artem Korenev, Punit~Singh Koura, Marie{-}Anne Lachaux, Thibaut Lavril, Jenya Lee, Diana Liskovich, Yinghai Lu, Yuning Mao, Xavier Martinet, Todor Mihaylov, Pushkar Mishra, Igor Molybog, Yixin Nie, Andrew Poulton, Jeremy Reizenstein, Rashi Rungta, Kalyan Saladi, Alan Schelten, Ruan Silva, Eric~Michael Smith, Ranjan Subramanian, Xiaoqing~Ellen Tan, Binh Tang, Ross Taylor, Adina Williams, Jian~Xiang Kuan, Puxin Xu, Zheng Yan, Iliyan Zarov, Yuchen Zhang, Angela Fan, Melanie Kambadur, Sharan Narang, Aur{\'{e}}lien Rodriguez, Robert Stojnic, Sergey Edunov,
  and Thomas Scialom. 2023.
\newblock \href {https://doi.org/10.48550/ARXIV.2307.09288} {Llama 2: Open foundation and fine-tuned chat models}.
\newblock \emph{CoRR}, abs/2307.09288.

\bibitem[{Treisman(1988)}]{doi:10.1080/02724988843000104}
Anne Treisman. 1988.
\newblock \href {https://doi.org/10.1080/02724988843000104} {Features and objects: The fourteenth bartlett memorial lecture}.
\newblock \emph{The Quarterly Journal of Experimental Psychology Section A}, 40(2):201--237.
\newblock PMID: 3406448.

\bibitem[{Vaswani et~al.(2017)Vaswani, Shazeer, Parmar, Uszkoreit, Jones, Gomez, Kaiser, and Polosukhin}]{DBLP:conf/nips/VaswaniSPUJGKP17}
Ashish Vaswani, Noam Shazeer, Niki Parmar, Jakob Uszkoreit, Llion Jones, Aidan~N. Gomez, Lukasz Kaiser, and Illia Polosukhin. 2017.
\newblock \href {https://proceedings.neurips.cc/paper/2017/hash/3f5ee243547dee91fbd053c1c4a845aa-Abstract.html} {Attention is all you need}.
\newblock In \emph{Advances in Neural Information Processing Systems 30: Annual Conference on Neural Information Processing Systems 2017, December 4-9, 2017, Long Beach, CA, {USA}}, pages 5998--6008.

\bibitem[{Wang et~al.(2019)Wang, Liu, Lin, Lin, and Han}]{DBLP:conf/cvpr/WangLLLH19}
Kuan Wang, Zhijian Liu, Yujun Lin, Ji~Lin, and Song Han. 2019.
\newblock {HAQ:} hardware-aware automated quantization with mixed precision.
\newblock In \emph{{CVPR}}, pages 8612--8620. Computer Vision Foundation / {IEEE}.

\bibitem[{Wang et~al.(2020)Wang, Wei, Dong, Bao, Yang, and Zhou}]{DBLP:conf/nips/WangW0B0020}
Wenhui Wang, Furu Wei, Li~Dong, Hangbo Bao, Nan Yang, and Ming Zhou. 2020.
\newblock Minilm: Deep self-attention distillation for task-agnostic compression of pre-trained transformers.
\newblock In \emph{NeurIPS}.

\bibitem[{Yu et~al.(2024)Yu, Yang, Li, Wang, Lin, Liu, Wang, and Wang}]{DBLP:conf/icml/YuYLWL0WW24}
Weihao Yu, Zhengyuan Yang, Linjie Li, Jianfeng Wang, Kevin Lin, Zicheng Liu, Xinchao Wang, and Lijuan Wang. 2024.
\newblock Mm-vet: Evaluating large multimodal models for integrated capabilities.
\newblock In \emph{{ICML}}. OpenReview.net.

\bibitem[{Zhu et~al.(2024)Zhu, Chen, Shen, Li, and Elhoseiny}]{DBLP:conf/iclr/Zhu0SLE24}
Deyao Zhu, Jun Chen, Xiaoqian Shen, Xiang Li, and Mohamed Elhoseiny. 2024.
\newblock Minigpt-4: Enhancing vision-language understanding with advanced large language models.
\newblock In \emph{{ICLR}}. OpenReview.net.

\end{thebibliography}

\appendix

\section{Benchmark Datasets Details}\label{app:benchmarks}

VQAv2 \citep{DBLP:conf/cvpr/GoyalKSBP17} includes approximately 11k test samples and focuses on visual question answering, where models must answer questions based on images depicting various real-world scenes. GQA \citep{DBLP:conf/cvpr/HudsonM19}, with around 12k test samples, emphasizes compositional reasoning through graph-structured annotations, assessing a model's ability to understand object relationships. VisWiz \citep{DBLP:conf/cvpr/Gurari0SGLGLB18}, containing 8k test samples, presents accessibility challenges with real-world images from visually impaired users, which are often of low quality and ambiguous, demanding robust model interpretation. SciQA-IMG \citep{DBLP:conf/nips/LuMX0CZTCK22} consists of around 4k test samples, targeting science-related visual question answering in specific domains. TextVQA \citep{DBLP:conf/cvpr/SinghNSJCBPR19}, with 5k test samples, focuses on understanding and answering questions from textual images. POPE \citep{DBLP:conf/emnlp/LiDZWZW23} contains approximately 9k test samples on three subsets: random, common, and adversarial. It evaluates a model's ability to predict human preference judgments on hallucination of multimodal tasks. MMBench (en) \citep{DBLP:journals/corr/abs-2307-06281}, with around 4k test samples, serves as a comprehensive benchmark for evaluating general-purpose multimodal models across various tasks, while MMBench (cn) \citep{DBLP:journals/corr/abs-2307-06281} is its Chinese translation. LLaVA-Wild \citep{DBLP:conf/nips/LiuLWL23a} has 60 test samples and emphasizes answering questions about complex, in-the-wild images. Finally, MM-Vet \citep{DBLP:conf/icml/YuYLWL0WW24} includes 218 samples and is designed to test multimodal capabilities across multiple visual and language tasks, providing a robust evaluation framework for emerging multimodal systems.

\section{Detailed TFLOPs of LVPruning with Different Vision Token Kept Ratio}\label{app:ratio_flops}

\begin{figure}[t]
    \centering
    \includegraphics[width=0.7\linewidth]{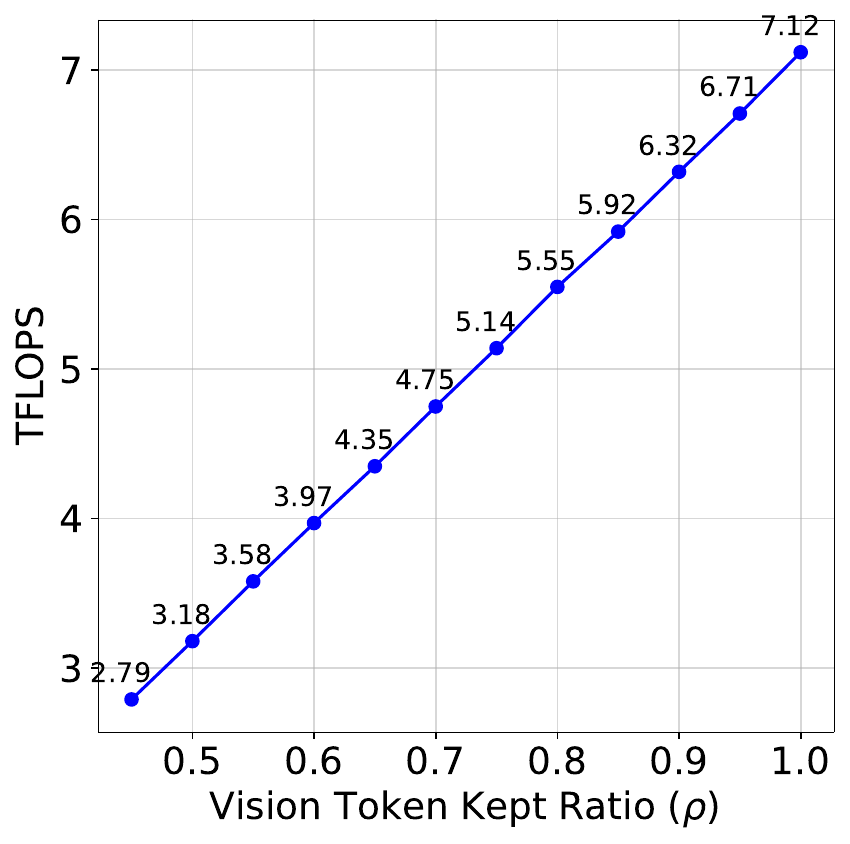}
    \caption{TFLOPs of LVPruning with different values of token kept ratio $\rho$ on LLaVA-1.5-7B.}
    \label{fig:d-flops}
\end{figure}

Figure \ref{fig:d-flops} shows the detailed TFLOPs of LVPruning with 0.05 change step of $\rho$. Additionally, compared with LLaVA-1.5-7B \citep{DBLP:journals/corr/abs-2310-03744}, the extra computation cost introduced by the inserted decision modules is 0.71 TFLOPs.

\end{document}